\documentclass[10pt,twocolumn,letterpaper]{article}

\usepackage{cvpr}
\usepackage{times}
\usepackage{epsfig}
\usepackage{graphicx}
\usepackage{amsmath}
\usepackage{amssymb}
\usepackage{amsmath}
\usepackage{amssymb}
\usepackage{diagbox}
\usepackage{algorithm}
\usepackage{algorithmicx}
\usepackage{algpseudocode}
\usepackage{graphics}
\usepackage{threeparttable}
\usepackage{color}
\usepackage[normalem]{ulem}
\usepackage{multirow}
\usepackage{float}
\usepackage{amsfonts}
\usepackage{bm}
\usepackage{array}
\usepackage[table]{xcolor}
\usepackage{colortbl}
\usepackage{url}
\usepackage{authblk}
\definecolor{mygray}{gray}{.8}
\makeatletter
\newcommand{\thickhline}{%
    \noalign {\ifnum 0=`}\fi \hrule height 1pt
    \futurelet \reserved@a \@xhline
}
\makeatother

\newcommand{\tabincell}[2]{\begin{tabular}{@{}#1@{}}#2\end{tabular}}

\usepackage[pagebackref=true,colorlinks,bookmarks=false]{hyperref}

\cvprfinalcopy 
\def\cvprPaperID{0353} 

\def\ourdataset{DHF1K}
\ifcvprfinal\fi
\begin{document}
\title{Revisiting Video Saliency: A Large-scale Benchmark and a New Model}
\author{Wenguan Wang $^1$, Jianbing Shen\thanks{Corresponding author: \textit{Jianbing Shen}. This work was supported in part by the Beijing Natural Science Foundation under Grant 4182056, the National Basic Research Program of China under Grant 2013CB328805,  the Fok Ying Tung Education Foundation under Grant 141067, and the Specialized Fund for Joint Building Program of Beijing Municipal Education Commission.}~~$^1$, Fang Guo $^1$, Ming-Ming Cheng $^2$, Ali Borji $^3$\\
$^1$\small Beijing Lab of Intelligent Information Technology, School of Computer Science, Beijing Institute of Technology, China\\
$^2$\small CCCE, Nankai University, China~~~$^3$\small Department of Computer Science, University of Central Florida, USA\\
{\tt\small wenguanwang.ai@gmail.com, \{shenjianbing, guofang\}@bit.edu.cn}\\
{\tt\small cmm@nankai.edu.cn, aborji@crcv.ucf.edu
}\\
{\tt\url{https://github.com/wenguanwang/DHF1K}}
}
\maketitle
\thispagestyle{empty}

\begin{abstract}
In this work, we contribute to video saliency research in two ways. First, we introduce a new benchmark for predicting human eye movements during dynamic scene free-viewing, which is long-time urged in this field.
Our dataset, named \ourdataset~(Dynamic Human Fixation), consists of 1K high-quality, elaborately selected video sequences 
spanning a large range of scenes, motions, object types and background complexity. Existing video saliency datasets lack variety and generality of common dynamic scenes and fall short in covering challenging situations in unconstrained environments. In contrast, \ourdataset~makes a significant leap in terms of scalability, diversity and difficulty, and is expected to boost video saliency modeling. Second, we propose a novel video saliency model that augments the CNN-LSTM network architecture with an attention mechanism to enable fast, end-to-end saliency learning. The attention mechanism explicitly encodes static saliency information, thus allowing LSTM to focus on learning more flexible temporal saliency representation across successive frames. Such a design fully leverages existing large-scale static fixation datasets, avoids overfitting, and significantly improves training efficiency and testing performance.
We thoroughly examine the performance of our model, with respect to state-of-the-art saliency models, on three large-scale datasets (\textit{i.e.}, \ourdataset, Hollywood2, UCF sports). Experimental results over more than 1.2K testing videos containing 400K frames demonstrate that our model outperforms other competitors.

\end{abstract}

\section{Introduction}
\label{sec:introduction}
Human visual system (HVS) has an astonishing ability to quickly select
visually important regions in its visual field. This cognitive
process enables humans to easily interpret complex scenes in
real time. Over the last few decades, several computational models have been proposed for imitating attentional mechanisms of HVS during static scene viewing. Significant advances have been achieved recently with the rapid spread of deep learning techniques and the availability of large-scale static gaze datasets (\textit{e.g.}, SALICON \cite{jiang2015salicon}).
In stark contrast, predicting observers' fixations during dynamic scene free-viewing has less been explored. This task, referred to as \textit{dynamic fixation prediction} or \textit{video saliency detection} is very useful for understanding human attentional behaviors and has several practical real-word applications (\textit{e.g.}, video captioning, compression, question answering, object segmentation, \textit{etc}). It is thus highly desired to have a standard, high-quality dataset composed of diverse and representative video stimuli.
Exiting datasets are severely limited in their coverage
and scalability, and they only include special scenarios such as limited human activities in constrained situations. None of them includes general, representative, and diverse instances in unconstrained, task-independent scenarios.
As a consequence, existing datasets often fail to offer
a rich set of fixations for learning video saliency and to assess models. Moreover, the existing datasets did not provide an evaluation server with standalone held out test set to avoid potential dataset over-fitting, which hinders further development on this topic.

While saliency benchmarks (\textit{e.g.}, MIT300 \cite{Judd_2012} and LSUN \cite{yu15lsun}) have been very instrumental in progressing the static saliency field, such standard widespread benchmarks are missing for video saliency modeling. We believe such benchmarks are highly needed to move the field forward. To this end, we propose a new benchmark ``\ourdataset~(Dynamic Human Fixation 1K)'' with a public server for reporting evaluation results on a preserved test set. Our benchmark contains a dataset that is unique in terms of generality, diversity and difficulty. It includes 1K videos with more than 600K frames and per-frame fixation annotations from 17 observers. The sequences have been carefully collected to include diverse scenes, motion patterns, object categories, and activities. \ourdataset~is accompanied with a comprehensive evaluation of several state-of-the-art approaches \cite{guo2010novel,seo2009static,rudoy2013learning,hou2009dynamic,fang2014video,hossein2015many,leboran2017dynamic,jiang2017predicting,bak2016two,itti1998model,harel2007graph,huang2015salicon,wang2017deep,pan2016shallow}. Moreover, each video is annotated with a main category label (\textit{e.g.}, daily activities, animals) and rich attributes (\textit{e.g.}, camera/content movement, scene lighting, presence of humans), which would enable a deeper understanding of gaze guidance in free viewing of dynamic scenes.


Further, we propose a novel CNN-LSTM architecture \cite{donahue2015long,noh2016image} based video saliency model with a supervised attention mechanism. CNN layers are utilized for extracting static features within input frames, while convolutional LSTM (convLSTM) \cite{xingjian2015convolutional} is utilized for sequential fixation prediction over successive frames. An attention module,
learned from existing large-scale image saliency datasets, is used to enhance spatially informative features of the CNN. Such a design helps disentangle underlying spatial and temporal factors of dynamic attention and allows convLSTM to learn temporal saliency representations efficiently. 

Our contributions are three-fold. \textbf{First}, we introduce a standard benchmark of 1K videos covering a wide range of scenes, motions, activities, \textit{etc}. \!To the best of our knowledge, the proposed dataset is the largest eye-tracking dataset for \textit{dynamic}, \textit{free-viewing} fixation prediction. \textbf{Second}, we present a novel attentive CNN-LSTM architecture for predicting human gaze in dynamic scenes, which explicitly encodes static attention into dynamic saliency representation learning by leveraging both static and dynamic fixation data. 
\textbf{Third}, we present a comprehensive analysis of video saliency models (the first one, to the best of our knowledge) on existing datasets (Hollywood-2, \!UCF sports), and our new \ourdataset~dataset. Results show that our model significantly outperforms previous methods.

\section{Related Work}
\label{sec:relatedwork}
\subsection{Video Eye-Tracking Datasets}
\label{sec:previousdatasets}
\vspace{-5pt}
There exist several datasets \cite{mathe2015actions,mital2011clustering,itti2004automatic,hadizadeh2012eye} for dynamic visual saliency prediction, but they are limited and often lack variety, generality and scalability of instances. Some statistics of these datasets are summarized in Table \ref{table}. 

The \textbf{Hollywood-2} dataset \cite{mathe2015actions} comprises all the $1, 707$ videos from Hollywood-2 action recognition dataset \cite{marszalek2009actions}. The videos are collected from $69$ Hollywood movies with $12$ action categories, such as eating, kissing and running. The human fixation data were tracked from $19$ observers belonging to $3$ groups for free viewing ($3$ observers), action recognition ($12$ observers), and context recognition ($4$ observers). Although this dataset is large, its content is limited to human actions and movie scenes. It mainly focuses on task-driven viewing mode, rather than free viewing. With $1, 000$ frames randomly sampled from Hollywood-2, we found $84.5$\% fixations are located around the faces.

The \textbf{UCF sports} fixation dataset \cite{mathe2015actions} contains $150$ videos taken from the UCF sports action dataset \cite{rodriguez2008action}. The videos cover $9$ common sports action classes, such as diving, swinging and walking. Similar to Hollywood-2, the viewers have been biased towards task-aware observation by being instructed to ``identify the actions occurring in the video sequence''. From the statistics of $1, 000$ frames randomly selected from UCF sports, we found $82.3$\% fixations fall inside the human body area.

The \textbf{DIEM} dataset \cite{mital2011clustering} is a public video eye-tracking dataset that has $84$ videos collected from publicly accessible video resources (\textit{e.g.}, advertisements, documentaries, sport events, and movie trailers, \textit{etc}). For each video, free-viewing fixations of around $50$ observers were collected. This dataset is mainly limited in its coverage and scale.

\textbf{Other datasets} are either limited in terms of variety and scale of video stimuli \cite{itti2004automatic,hadizadeh2012eye}, or collected for a special purpose (\textit{e.g.}, salient objects in videos \cite{wang2015consistent}). More importantly, none of the aforementioned datasets includes a preserved test set for avoiding potential data overfitting, which has seriously hampered the research process.

\begin{table}
\centering
\resizebox{0.49\textwidth}{!}{
\setlength\tabcolsep{2pt}
\renewcommand\arraystretch{1.0}
\begin{tabular}{r|c|c|c|c|c|c}  
\hline\thickhline
Dataset~~~~~ &Year &Videos &Resolution &Duration(s) &Viewers &Task\\
\hline
\hline
CRCNS \cite{itti2004automatic} &2004 &50   &$640\times480$ &6-94 &15 &task-goal\\
Hollywood-2 \cite{mathe2015actions} &2012  &1,707    &$720\times480$ &2-120&19 &task-goal\\
UCF sports \cite{mathe2015actions} &2012  &150     &$720\times480$ &2-14 &19 &task-goal\\
DIEM \cite{mital2011clustering} &2011 &84   &$1280\times720$ &27-217 &$\sim$50 &free-view\\
SFU \cite{hadizadeh2012eye} &2012 &12   &$352\times288$ &3-10 &15 &free-view\\
\ourdataset (\textbf{Ours})     &2017   &1,000  &$640\times360$ &17-42 &17 &free-view\\
\hline
\end{tabular}
}
\vspace*{0pt}
\caption{\textbf{Statistics of typical dynamic eye-tracking datasets.} }\label{table}
\vspace{-10pt}
\end{table}

\subsection{Computational Models for Fixation Prediction}
\label{sec:saliencymodels}
\vspace{-5pt}
The study of human gaze patterns in static scenes has received significant interests, which can be dated back to \cite{itti1998model,itti2001computational}.
\textbf{Early static saliency models} \cite{le2006coherent,zhang2008sun,gao2005discriminant,bruce2006saliency,harel2007graph,hou2007saliency,judd2009learning,wang2017stereoscopic}
are mostly based on the \textit{contrast} assumption that conspicuous visual
features ``pop-out'' and involuntarily capture attention (see \cite{borji2013state,borji2013quantitative} for review). Computational models compute multiple visual features such as color, edge, and orientation at multiple spatial scales to produce a ``saliency map": an image distribution predicting the conspicuity of specific locations and their likelihood in attracting attention \cite{itti2001computational,mital2011clustering}. The locations with more distinct feature responses over surroundings usually gain higher saliency values. \textbf{Deep learning based static saliency models} \cite{vig2014large,kruthiventi2017deepfix,huang2015salicon,liu2016learning,pan2016shallow,jetley2016end,wang2017deepcrop,wang2017deep} have achieved astonishing improvements, relying on the powerful end-to-end learning ability of neural network and the availability of large-scale static saliency datasets \cite{jiang2015salicon}.

\textbf{Previous investigations of dynamic human fixation} \cite{gao2008discriminant,guo2010novel,mahadevan2010spatiotemporal,rudoy2013learning,seo2009static,hou2009dynamic,fang2014video,hossein2015many, leboran2017dynamic} leveraged both static stimulus features and temporal information (\textit{e.g.}, optical flow, difference-over-time, \textit{etc}). Some of those studies \cite{gao2008discriminant,mahadevan2010spatiotemporal,seo2009static} can be viewed as extensions of exiting static saliency models with additional motion features. Those models are mainly bound to significant feature engineering and limited representation ability of hand-crafted features. To date, only a few \textbf{deep learning based video saliency models} \cite{bak2016two,jiang2017predicting} exist in this field. They are mainly based on 
two-stream network architecture \cite{bak2016two} that accounts for color images and motion fields separately, or two-layer LSTM with object information \cite{jiang2017predicting}. These works show a better performance and demonstrate the potential advantages in applying neural networks to this problem. However, they do not 1) consider attentive mechanisms; 2) utilize existing large-scale static fixation datasets; and 3) exhaustively assess their performance over large amount of data.

There are some \textbf{salient object detection models} \cite{Liu2007,achanta2009frequency,ChengPAMI,wang2016correspondence,wang2015saliency,wang2017video,SalObjBenchmark,wang2018saliencyPAMI,DSSalCVPR2017} that attempt to uniformly highlight salient object regions in images or videos. Those models are often task-driven and focus on inferring the main object, instead of investigating the behavior of the HVS during scene free viewing.

\subsection{Attention Mechanisms in Neural Networks}
\label{sec:attentionnetwork}
Recently, incorporating attention mechanisms into network architectures has shown great success in several computer vision \cite{yang2016stacked,cao2015look,wangresidual} and natural language processing tasks \cite{rush2015neural,parikh2016decomposable}. In such studies, attention is learned in an automatic, top-down, and task-specific manner, allowing the network to focus on the most relevant parts in images or sentences. In this paper, we use attention for enhancing intra-frame salient features, thus allowing the LSTM to model dynamic representations more easily. In contrast to previous models learning attentions implicitly, our attention module encodes strong static saliency information and can be learned from existing static saliency dataset in a supervised manner. This design leads to improved generality and prediction performance. It is the first attempt to incorporate a supervised attention mechanism into the network structure to achieve state-of-art results in dynamic fixation prediction.

\section{\ourdataset~Dataset}
\label{sec:dataset}
\begin{figure}
  \centering
      \includegraphics[width=0.99 \linewidth]{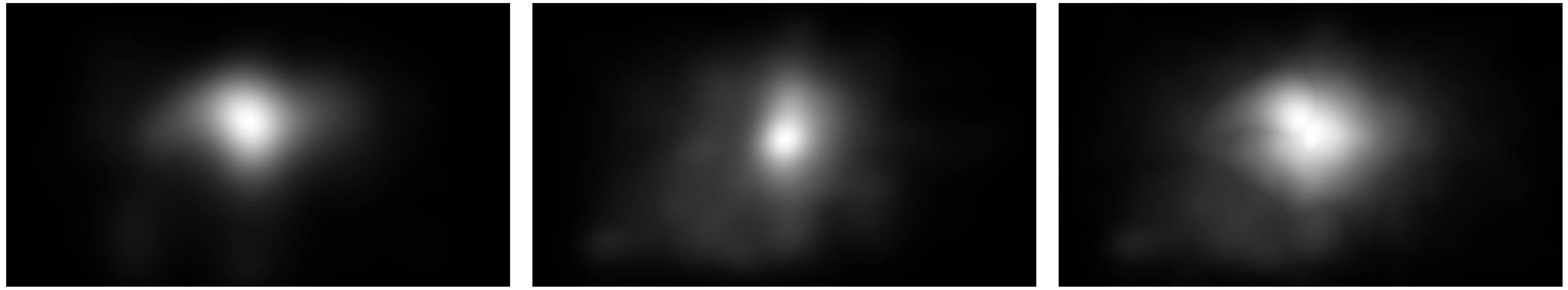}
     \\
     \mbox{}\hfill \footnotesize (a) \hfill\mbox{}
     \mbox{}\hfill \footnotesize(b) \hfill\mbox{}
     \mbox{}\hfill \footnotesize(c) \hfill\mbox{}
\caption{\textbf{Average annotation maps of three datasets} used in benchmarking: (a) Hollywood-2, (b) UCF sports, (c) \ourdataset.}
\label{fig1}
\vspace{-10pt}
\end{figure}

\begin{figure*}
  \centering
      \includegraphics[width=0.99 \linewidth]{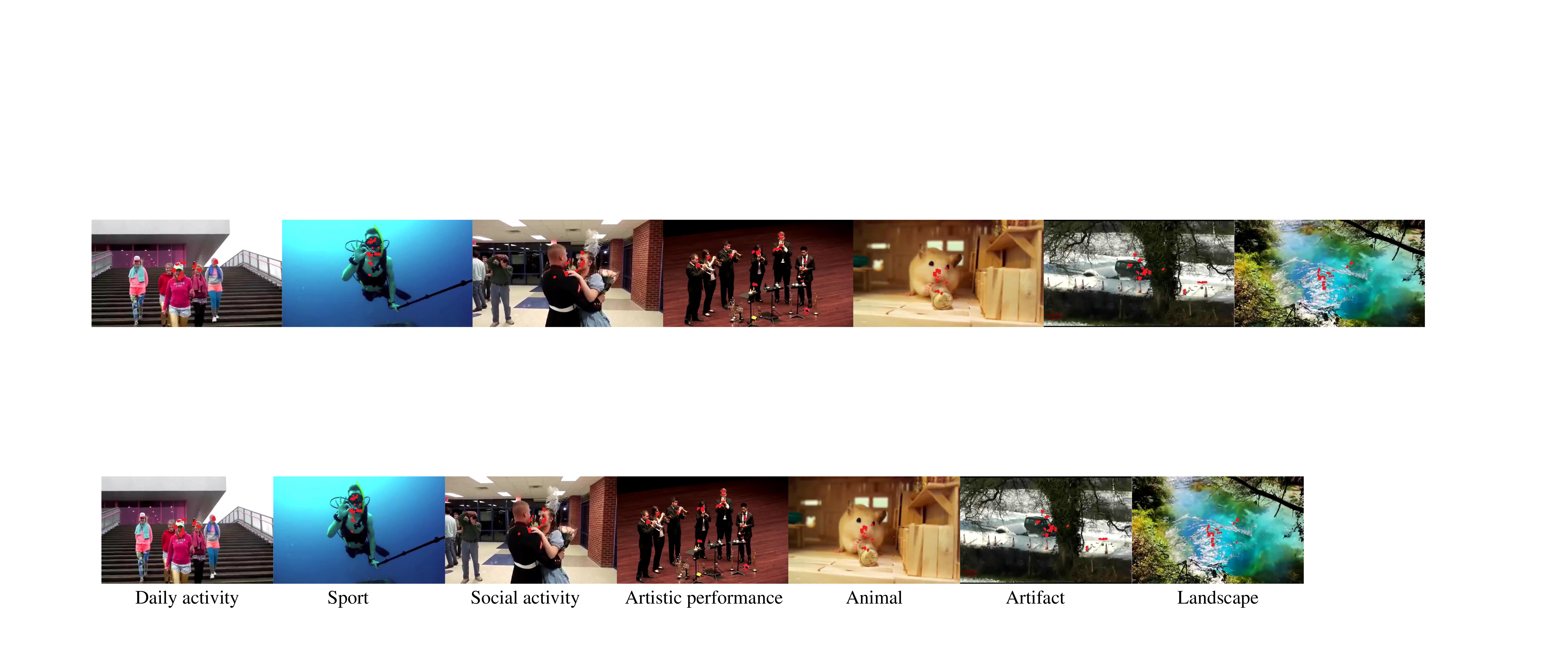}
\caption{\textbf{Example frames from \ourdataset}~ with fixations (red dots) and corresponding categories.}
\label{fig2}
\vspace{-15pt}
\end{figure*}

\begin{table}
\centering
\begin{threeparttable}
\resizebox{0.49\textwidth}{!}{
\setlength\tabcolsep{2pt}
\begin{tabular}{c||c|c|c|c|c|c|c}  
\hline\thickhline
\multirow{2}*{\ourdataset}
&\multicolumn{4}{c|}{Human}
&\multirow{2}*{Animal}
&\multirow{2}*{Artifact}
&\multirow{2}*{Scenery}
\\
\cline{2-5}
 &Daily ac. &Sports &Social ac. &Art & &\\
\hline
\hline
\#sub-classes\tnote{*}~ &20  &29    &13 &10 &36 &21 &21\\
\#videos    &134  &185    &116 &101 &192 &162 &110\\
\hline
\end{tabular}
}
\begin{tablenotes}
\footnotesize
\item$^*$Number of sub-classes in each category is reported. For example,\\
\textit{Sports} has sub-classes like \textit{swimming}, \textit{jumping}, \textit{etc}.
\end{tablenotes}
\end{threeparttable}
\vspace{-2pt}
\caption{\textbf{Statistics for video categories} in \ourdataset~dataset.}\label{table1}
\vspace{-10pt}
\end{table}

We introduce \ourdataset, a large-scale dataset of gaze in free-viewing of videos. Our dataset includes 1K videos with diverse content and length, with eye-tracking annotations from $17$ observers. Fig. \ref{fig1} shows the center bias of \ourdataset, compared to Hollywood-2, and UCF sports datasets.

\textbf{Stimuli.} The collection of dynamic stimuli mainly follows the following 4 principles.

\noindent{\small\textbullet}~\textit{Large scale and high quality}. Both scale and quality are necessary to ensure the content diversity of a dataset and is crucial to guarantee a longer lifespan for a benchmark. 
To this end, we searched the Youtube engine with about 200 key terms (\textit{e.g.,} dog, walking, car, \textit{etc}) and carefully selected $1, 000$ video sequences from the retrieval results. All videos were converted from their original sources to a 30 fps Xvid MPEG-4 video file in an AVI container and were resized uniformly into $640\times360$ spatial resolution. Thus, \ourdataset~comprises a total $1, 000$ video sequences with $582, 605$ frames with total duration of $19, 420$ seconds.

\begin{table}
\centering
\resizebox{0.49\textwidth}{!}{
\setlength\tabcolsep{4pt}
\begin{tabular}{c||c|c|c|c|c|c|c|c|c|c}  
\hline\thickhline
\multirow{2}*{\ourdataset}
&\multicolumn{3}{c|}{Content motion}
&\multicolumn{3}{c|}{Camera motion}
&\multicolumn{4}{c}{\#Objects}
\\
\cline{2-11}
         &stable &slow &fast &stable &slow &fast &0 &1 &2 &$\geq$3\\
\hline
\hline
\#videos &126 &505 &369    &343 &386 &271 &56 & 335 &254 &355\\
\hline
\end{tabular}
}
\vspace{0pt}
\caption{\textbf{Statistics regarding motion patterns and number of main objects} in \ourdataset~dataset.}\label{table2}
\vspace{-6pt}
\end{table}

\begin{table}
\centering
\resizebox{0.38\textwidth}{!}{
\setlength\tabcolsep{4pt}
\begin{tabular}{c||c|c|c|c|c|c|c}  
\hline\thickhline
\multirow{2}*{\ourdataset}
&\multicolumn{3}{c|}{Scene illumination}
&\multicolumn{4}{c}{\#People}
\\
\cline{2-8}
         &day &night &indoor  &0 &1 &2 &$\geq$3\\
\hline
\hline
\#videos &577 &37 &386     &345 & 307 &236 &112\\
\hline
\end{tabular}
}
\vspace{4pt}
\caption{\textbf{Statistics regarding scene illumination and number of people} in \ourdataset~dataset.}\label{table3}
\vspace{-10pt}
\end{table}

\noindent{\small\textbullet}~\textit{Diverse content}. Stimulus diversity is essential for avoiding overfitting and to delay performance saturation. It offers evenly distributed exogenous control for studying person-external stimulus factors during scene free-viewing. In \ourdataset, each video is manually annotated with a category label (totally $150$ classes). Those labels are further classified into $7$ main categories (see Table \ref{table1}). 
Those semantic annotations would enable a deeper understanding of the high-level stimuli factors guiding human gaze in dynamic scenes and be indicative for potential research. In Fig. \ref{fig2}, we show example frames from each category.

\noindent{\small\textbullet}~\textit{Varied motion patterns}. Previous investigations \cite{itti2005quantifying,gao2008discriminant,mital2011clustering} suggested that motion is one of the key factors that directs attention allocation in dynamic viewing. For this, \ourdataset~ is designed to span varied motion patterns (\textit{stable-/slow-/fast-motion} of content and camera). Please see Table \ref{table2} for the information regarding motion patterns.

\noindent{\small\textbullet}~\textit{Various objects}. Previous studies \cite{wolfe2007guided,li2014secrets,borji2015salientTip} in cognitive and computer vision confirmed that object information is indicative to human fixations. The objects in the dataset vary in their type (\textit{e.g.}, \textit{human}, \textit{animal}, in Table \ref{table1}) and frequency (see Table \ref{table2}). For each video, five subjects were instructed to count the number of the main objects. The majority vote of their counts was considered as the final count. 

For completeness, in Table \ref{table3}, we offer the information of the scene illumination and the amount of humans in the dataset. As demonstrated in \cite{motter1994neural}, luminance is an important exogenous factor for attentive selection. Further, human beings are important high-level stimuli \cite{cerf2008predicting,bylinskii2016should} in free-viewing.

\textbf{Apparatus and technical specifications.}
Participants' eye movements were monitored binocularly using a Senso Motoric Instruments (SMI) RED 250 system at a sampling rate of 250 Hz.
The dynamic stimuli were displayed on a 19'' display (resolution $1440\times900$). A headrest was used to help participants' heads still at a distance of around $68$ cm, as advised by the product manual.

\textbf{Participants.} $17$ participants ($10$ males and $7$ females, aging between $20$ and $28$) who passed
the calibration of the eye tracker and had less than $10$\% fixation dropping rate, were quantified for our eye tracking
experiment. 
All participants had normal or corrected-to-normal vision. All subjects had not 
seen the stimuli in \ourdataset~before. All provided informed consent and were na\"{\i}ve to the underlying purposes of the experiment.

\textbf{Data capturing.} 
The stimuli were equally  partitioned into 10 non-overlapping sessions. Participants were required to freeview 10 sessions of videos in random order. In each session, the videos were also displayed at random. Before the experiments, every participant was calibrated using the standard routine in product manual with
recommended settings for the best results. 
To avoid eye fatigue, each video presentation was followed by a 5-second
waiting interval with black screen. After undergoing
a session of videos, the participant can take a rest until
she was ready for viewing the next session. 
Finally, $51,038,600$ fixations were recorded from 17 subjects on $1,000$ videos.

\textbf{Training/testing split.}
We split $1,000$ dynamic stimuli into separate training, validation and test sets. Following random selection, we arrive at a
unique split consisting of $600$ training and $100$ validation videos with publicly available fixation records, as well as $300$ test videos with annotations held-out for benchmarking purpose.
\section{Our Approach}
\label{sec:ourapproach}
\begin{figure*}
  \centering
      \includegraphics[width=\linewidth]{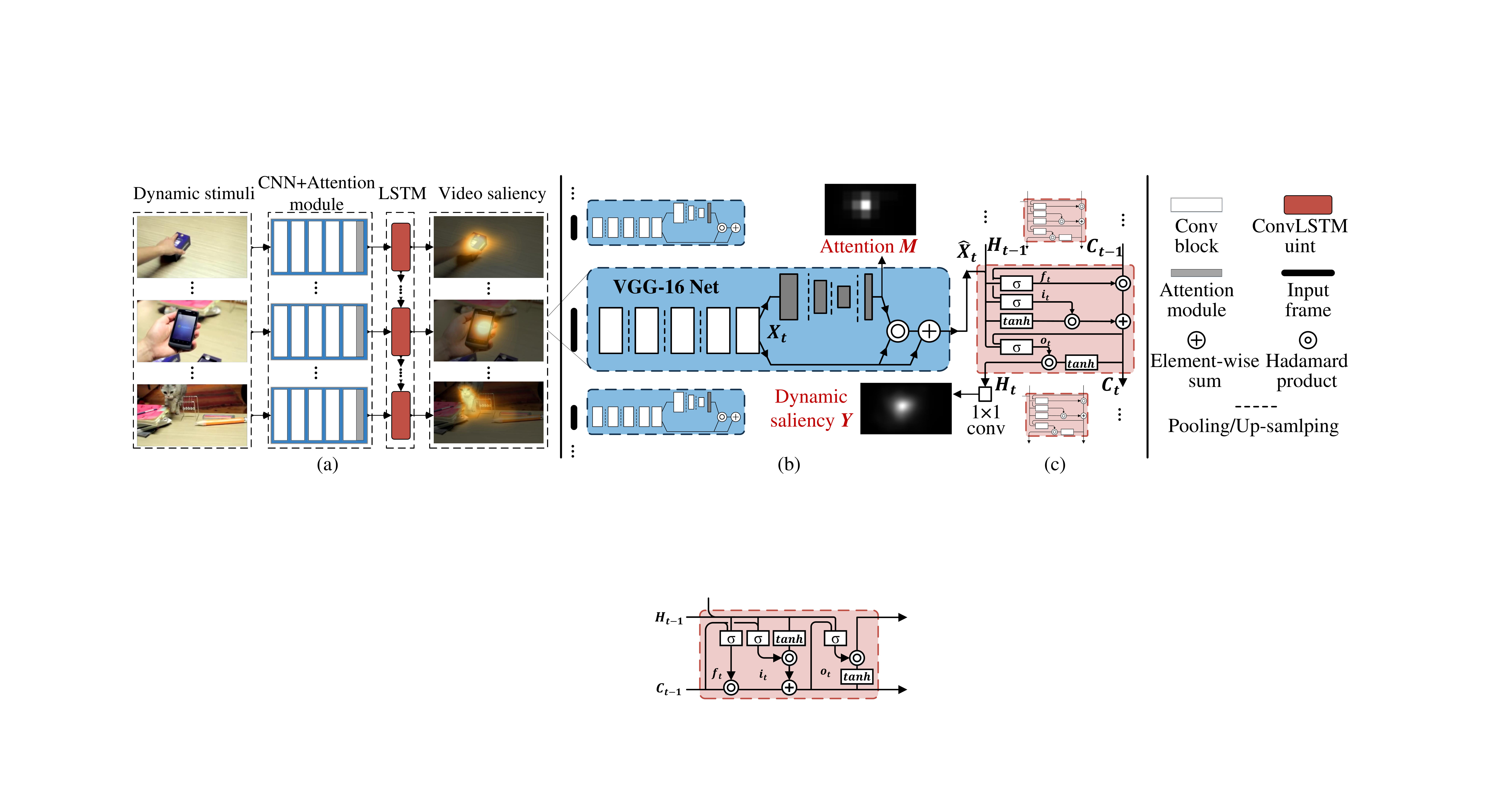}
\caption{\textbf{Network architecture of the proposed video saliency model.} (a) Attentive CNN-LSTM architecture. (b) CNN layers with attention module are used for learning intra-frame static features, where the attention module is learned with the supervision from static saliency data. (c) ConvLSTM used for learning sequential saliency representations. }
\label{fig3}
\vspace{-12pt}
\end{figure*}

\textbf{Overview.} Fig. \ref{fig3} presents the overall architecture of our video saliency model. It is based on a CNN-LSTM architecture that combines convolutional network and recurrent model to exploit both spatial and temporal information for predicting video saliency. The CNN-LSTM network is extended with a supervised attention mechanism, which explicitly captures static saliency information and allows the LSTM to focus on learning dynamic information. 
The attention module is trained from rich static eye-tracking data. Thus our model is able to produce accurate, spatiotemporal saliency with improved generalization ability. Next, we explain each component of our model in detail.

\textbf{CNN-LSTM architecture.} Formally, given an input video $\{I_t\}_t$, we first obtain a sequence of convolutional features $\{\mathcal{X}_t\}_t$ from CNN. Then the features $\{\mathcal{X}_t\}_t$ are fed into a convLSTM \cite{xingjian2015convolutional} as input. Here, the convLSTM is used for modeling the temporal dynamic nature of this sequential problem, which is achieved by incorporating memory units with gated operations. Additionally, through replacing dot products with convolutional operations, the convLSTM is able to preserve spatial information, which is essential for making spatially-variant pixel-level prediction.

More specifically, the convLSTM utilizes three convolution gates (\textit{input}, \textit{output} and \textit{forget}) to control the flow of signal within the cell. 
With the input feature $\mathcal{X}_t$ at time step $t$, the convLSTM outputs a hidden state $\mathcal{H}_t$ and maintains a memory cell $\mathcal{C}_t$ for controlling state update and output:
\begin{small}
\begin{flalign}
    \!\!\!\!i_{t}\!&= \sigma(W^\mathcal{X}_i\!\ast\!\mathcal{X}_t\!+\!W^\mathcal{H}_i\!\ast\!\mathcal{H}_{t-1}\!+\!W^\mathcal{C}_i\circ \mathcal{C}_{t-1}\!+\!b_i),\\
    \!\!\!\!f_{t}\!&= \sigma(W^\mathcal{X}_f\!\ast\!\mathcal{X}_t\!+\!W^\mathcal{H}_f\!\ast\!\mathcal{H}_{t-1}\!+\!W^\mathcal{C}_f\circ \mathcal{C}_{t-1}\!+\!b_f),\\
    \!\!\!\!o_{t}\!&= \sigma(W^\mathcal{X}_o\!\ast\!\mathcal{X}_t\!+\!W^\mathcal{H}_o\!\ast\!\mathcal{H}_{t-1}\!+\!W^\mathcal{C}_o\circ \mathcal{C}_{t}\!+\!b_o),\\
    \!\!\!\!\mathcal{C}_{t}\!&= f_t \circ \mathcal{C}_{t-1}\!+\!i_t\!\circ\!\tanh(W^\mathcal{X}_c\!\ast\!\mathcal{X}_t\!+\!W^\mathcal{H}_c\!\ast\! \mathcal{H}_{t-1}\!+\!b_c),\\
    \!\!\!\!\mathcal{H}_{t}\!&= o_t\circ\tanh(\mathcal{C}_t),
    \label{eq:2}
\end{flalign}
\end{small}
$i_{t}$, $f_{t}$, $o_{t}$ are the gates. $\sigma$ and $\tanh$ are the activation functions of logistic sigmoid and hyperbolic tangent, respectively.
`$\ast$' denotes the convolution operator and `$\circ$' represents Hadamard product. 
The dynamic fixation map can be obtained via convolving the hidden states $\mathcal{H}$ with a $1\times1$ kernel (see Fig. \ref{fig3} (c)).

In our implementation, the first five conv blocks of VGG-16 \cite{simonyan2014very} are used. For preserving more spatial details, we remove \textit{pool4} and \textit{pool5} layers, which results in $\times8$ instead of $\times32$ downsampling. At time step $t$, with an input frame $I_t$ with 224$\times$224 resolution, we have $\mathcal{X}_t\in \mathbb{R}^{28\times28\times512}$ and a 28$\times$28 dynamic saliency map from the convLSTM. The kernel size of the conv layer in convLSTM is set as 3.

\textbf{Attention module.} We extend above CNN-LSTM architecture with an attention mechanism, which is learned from existing static fixation data in a  supervised manner. Such design is mainly driven by the following three motivations:

\noindent{\small\textbullet}~Previous studies \cite{itti2005quantifying, wischnewski2010look} shown that human attention is guided by both static and dynamic factors. Through the additional attention module, the CNN is enforced to generate a more explicit spatial saliency representation. This helps disentangle underlying spatial and temporal factors of dynamic attention, and allows convLSTM better capture temporal dynamics.

\noindent{\small\textbullet}~CNN-LSTM architecture introduces a large number of parameters for modeling spatial and temporal patterns. However, for sequential data such as videos, obtaining labelled data is costly. Even though there are large-scale datasets like \ourdataset~that have 1K videos, the amount of training data is still insufficient, considering the high correlation among frames within same video. The supervised attentive module is able to leverage existing rich static fixation data to improve the generalization power of our model. 

\noindent{\small\textbullet}~In VGG-16, we remove the last two pooling layers to obtain a large feature map. \!This dramatically decreases the receptive field ($212$$\times$$212$$\rightarrow$$140$$\times$$140$), which cannot cover the whole frame ($224$$\times$$224$). \!To remedy this, we insert a set of down- and up-sampling operations into the attention module, which would enhance the intra-frame saliency information with an enlarged receptive field. By this, our model is able to make more accurate predictions from a global view.

As demonstrated in Fig. \ref{fig3} (b), our attentive module is built upon the \textit{conv5-3} layer, as an additional branch of several conv layers interleaved with pooling, and upsampling operations. Given the input feature $\mathcal{X}$, with pooling layers, the attention module generates a downsampled attention map (7$\times$7) with an enlarged receptive field (260$\times$260). Then the small attention map is $\times4$ upsampled as the same spatial dimensions of $\mathcal{X}$. Let $M\in[0,1]^{28\times 28}$ be the upsampled attention map, the feature $\mathcal{X}\in\mathbb{R}^{28\times28\times512}$ from \textit{conv5-3} layer can be further enhanced by:
\vspace{-2pt}
\begin{equation}\small\begin{aligned}
    \hat{\mathcal{X}}^c = M\circ \mathcal{X}^c,
\end{aligned}\label{eq:eq2}\vspace{-2pt}\end{equation}
where $c\in \{1,\ldots,512\}$ is the index of the channel. Here, the attention module
work as a feature selector to enhance the feature representation.

The above attention module may lose useful information for learning a dynamic saliency representation, as the attention module only considers static saliency information in still video frames. For this, inspired by the recent advances of attention mechanism and residual connection \cite{he2016deep,wangresidual}, we improve Eq. \ref{eq:eq2} in residual form:
\vspace{-2pt}
\begin{equation}\small\begin{aligned}
    \hat{\mathcal{X}}^c = (1+M)\circ \mathcal{X}^c.
\end{aligned}
\vspace{-2pt}
\label{eq:eq3}\end{equation}
With the residual connection, both the original CNN features and the enhanced features are combined and fed to the LSTM model. 
In \S \ref{sec:performance} and \S \ref{sec:ablation}, more detailed explorations for the attention module are offered.

Different from previous attention mechanisms that learn task-related attention in an implicit way, our attention module can learn from existing large-scale static fixation data in an explicit and supervised manner (detailed in next part).


\textbf{Loss function.} We use the following loss function \cite{huang2015salicon} that considers three different saliency
evaluation metrics instead of one. The rationale here is that no single metric can fully capture how satisfactory a saliency map is.

We denote the predicted saliency map as $Y\!\!\in\![0,1]^{28\times 28}$, the map of fixation locations as $P\!\!\in\!\!\{0,1\}^{28\times 28}$ and the continuous saliency map (distribution) as $Q\!\!\in\!\![0,1]^{28\times 28}$. Here the fixation map $P$ is discrete, that records whether a pixel receives human fixation. The continuous saliency map is obtained via blurring each fixation location with a small Gaussian kernel. Our loss functions is defined as follows:
\vspace{-2pt}
\begin{equation}\small
    \begin{aligned}
    \!\!\!\mathcal{L}(Y\!,\!P\!,\!Q)\!= &\mathcal{L}_{K\!L}(Y,\!Q)\!+\!\alpha_1\mathcal{L}_{CC}(Y,\!Q)\!+\!\alpha_2\mathcal{L}_{N\!S\!S}(Y,\!P),
    \end{aligned}
    \label{eq:3}
    \vspace{-2pt}
\end{equation}
where $\mathcal{L}_{KL}$, $\mathcal{L}_{CC}$ and $\mathcal{L}_{NSS}$ are the \textit{Kullback-Leibler (KL) divergence},
the \textit{Linear Correlation Coefficient (CC)}, and the \textit{Normalized Scanpath Saliency (NSS)}, respectively, which are
derived from commonly used metrics to evaluate saliency prediction models. $\alpha$s are balance parameters and are empirically set to $\alpha_1 = \alpha_2 = 0.1$.

$\mathcal{L}_{KL}$ is widely adopted for training saliency models and is chosen as the primary loss in our work:
\vspace{-2pt}
\begin{equation}\small
    \begin{aligned}
     \mathcal{L}_{KL}(Y, Q) = \sum\nolimits_{x}Q(x)\log\big(\frac{Q(x)}{Y(x)}\big).
    \end{aligned}
    \label{eq:4}
    \vspace{-2pt}
\end{equation}

$\mathcal{L}_{CC}$ measures the linear relationship between $Y$ and $Q$:
\vspace{-2pt}
\begin{equation}\small
    \begin{aligned}
     \mathcal{L}_{CC}(Y, Q) = -\frac{cov(Y,Q)}{\rho(Y)\rho(Q)},
    \end{aligned}
    \label{eq:5}
    \vspace{-2pt}
\end{equation}
where $cov(Y,Q)$ is the covariance of $Y$ and $Q$, and $\rho(\cdot)$ stands for standard deviation.

$\mathcal{L}_{NSS}$ is derived from NSS metric:
\vspace{-2pt}
\begin{equation}\small
    \begin{aligned}
     \mathcal{L}_{NSS}(Y, P) = -\frac{1}{N}\sum\nolimits_{x}\overline{Y}(x)\times P(x),
    \end{aligned}
    \label{eq:5}
    \vspace{-2pt}
\end{equation}
where $\overline{Y} = \frac{Y-\mu(Y)}{\rho(Y)}$ and $N = \sum_{x}P(x)$. It is calculated by taking the mean of scores from the normalized saliency map $\overline{Y}$ (with zero mean and unit standard deviation) at human eye fixations $P$. Since $CC$ and $NSS$ are similarity metrics, their negatives are adopted for minimization.

\textbf{Training protocol.} Our model is iteratively trained with sequential fixation and image data. In training, a video training batch is cascaded with an image training batch. More specifically, in a video training batch, 
we apply a loss defined over
the final dynamic saliency prediction from LSTM. Let $\{Y^{d}_t\}_{t=1}^{T}$, $\{P^{d}_t\}_{t=1}^{T}$, and $\{Q^{d}_t\}_{t=1}^{T}$ denote the dynamic saliency predictions, the dynamic fixation sequence and the continuous ground-truth saliency maps, we minimize the following loss:
\vspace{-2pt}
\begin{equation}\small
    \begin{aligned}
    \mathcal{L}^{d} = \sum\nolimits_{t=1}^T\mathcal{L}(Y^{d}_t,P^{d}_t,Q^{d}_t).
    \end{aligned}
    \label{eq:6}
    \vspace{-2pt}
\end{equation}
In this process, the attention module is trained in an implicit way, since we do not have the groundtruth fixation of each frame in static scene.

In an image training batch, we only train our attention module via minimizing:
\vspace{-2pt}
\begin{equation}\small
    \begin{aligned}
    \mathcal{L}^{s} = \mathcal{L}(M,P^{s},Q^{s}),
    \end{aligned}
    \label{eq:7}
    \vspace{-2pt}
\end{equation}
where the $M$, $P^{s}$, $Q^{s}$ indicate the attention map for our static attention module, the ground-truth static fixation map, and the ground-truth static saliency map.
In this process, the training of attention module is supervised by the ground-truth static fixation. Note that, in image training batch, we do not train our LSTM module, as it is used for learning the dynamic representation.

For each video training batch, 20 consecutive frames from the same video are used. Both the video and the start frame are randomly selected. For each image training batch, we set the batch size as 20, and the images are randomly sampled from existing static fixation dataset. More implementation details can be found in \S~\ref{sec:experimentsetup}.

\section{Experiments}
\label{sec:Experimentalresults}
\subsection{Experimental Setup}
\label{sec:experimentsetup}
\begin{table*}[t]
\centering
\begin{threeparttable}
\resizebox{\textwidth}{!}{
\setlength\tabcolsep{1pt}
\renewcommand\arraystretch{1.2}
\begin{tabular}{l|r|c|c|c|c|c||c|c|c|c|c||c|c|c|c|c}  
\hline\thickhline
\multirow{2}*{} &\multirow{2}*{\diagbox[height=2.80em,width=7.2em,trim=l]{~Method}{\!Dataset~}}
&\multicolumn{5}{c||}{\ourdataset} &\multicolumn{5}{c||}{Hollywood-2} &\multicolumn{5}{c}{UCF sports}\\
\cline{3-17}
&&AUC-J$\uparrow$ &~SIM$\uparrow$  &s-AUC$\uparrow$ &~~CC$\uparrow$~~ &~NSS$\uparrow$ &AUC-J$\uparrow$ &~SIM$\uparrow$  &s-AUC$\uparrow$ &~~CC$\uparrow$~~ &~NSS$\uparrow$ &AUC-J$\uparrow$ &~SIM$\uparrow$  &s-AUC$\uparrow$ &~~CC$\uparrow$~~ &~NSS$\uparrow$\\
\hline
\hline
&$^*$PQFT \cite{guo2010novel}                                           &0.699 	&0.139 	 	&0.562 	&0.137 	&0.749  &0.723 	&0.201 	 	&0.621 	&0.153 	&0.755 &0.825 	&0.250 	 	&0.722 	&0.338 	&1.780\\
&$^*$Seo \textit{et al}. \cite{seo2009static}                  &0.635 	    &0.142 	   &0.499 	 &0.070 	&0.334 &0.652 	    &0.155 	   &0.530 	 &0.076 	&0.346  &0.831 	    &0.308 	   &0.666 	 &0.336 	&1.690\\
Dynamic&$^*$Rudoy \textit{et al}. \cite{rudoy2013learning}             &0.769  	&0.214  	&0.501  	 	&0.285 	&1.498 &0.783  	&0.315  	&0.536  	 	&0.302 	&1.570 &0.763  	&0.271  	&0.637  	 	&0.344 	&1.619 \\
models&$^*$Hou \textit{et al}. \cite{hou2009dynamic}                      &0.726 	    &0.167	     	&0.545 	&0.150 	     &0.847 &0.731 	    &0.202	     	&0.580 	&0.146 	     &0.684 &0.819 	    &0.276	     	&0.674 	&0.292 	     &1.399\\
&$^*$Fang \textit{et al}. \cite{fang2014video}                      &0.819  	&0.198  	 	&0.537  	&0.273  	&1.539  &0.859  	&0.272  	 	&0.659 	&0.358 	&1.667 &0.845  	&0.307  	 	&0.674  	&0.395  	&1.787\\
&$^*$OBDL \cite{hossein2015many} &0.638  	&0.171 	 	  	&0.500  	&0.117  &0.495  	&0.640  	 	&0.170 	&0.541 	&0.106 &0.462 	&0.759 &0.193  	&0.634  	&0.234 &1.382\\
&$^*$AWS-D \cite{leboran2017dynamic}        &0.703	&0.157 	 	&0.513  	&0.174  	&0.940  &0.694  	&0.175  	 	&0.637 	& 0.146 	&0.742 &0.823 	&0.228  	 	&0.750  	&0.306  	&1.631\\
&OM-CNN \cite{jiang2017predicting}                                   &0.856  &0.256  	&0.583  	 	&0.344 	&1.911 &0.887  	&0.356  	&0.693  	 	&0.446 	&2.313 &0.870  	&0.321  	&0.691  	 	&0.405 	&2.089 \\
&Two-stream \cite{bak2016two}            &0.834 	 	&0.197 	&0.581 &0.325	&1.632 &0.863 	 	&0.276 	&0.710 &0.382	&1.748 &0.832  	&0.264  	&0.685  	 	&0.343 	&1.753 \\
\hline
\hline
&$^*$ITTI	 \cite{itti1998model}                            &0.774  	&0.162  &0.553  &0.233  &1.207 &0.788  	&0.221  	 	&0.607  	&0.257 	&1.076 &0.847  	&0.251  	 	&0.725  	&0.356  	&1.640 \\
& $^*$GBVS \cite{harel2007graph}                           &0.828  	&0.186  	 	&0.554  	&0.283  	&1.474 &0.837  	&0.257  	 	&0.633  	&0.308   	&1.336  &0.859  	&0.274  	 	&0.697  	&0.396  	&1.818 \\
Static&SALICON \cite{huang2015salicon}                      &0.857  	&0.232  	 	&0.590  	&0.327  	&1.901 &0.856  	&0.321  	 	&0.711  	&0.425  	&2.013  &0.848  	&0.304  	 	&0.738  	&0.375  	&1.838 \\
models&Shallow-Net \cite{pan2016shallow}   &0.833 	 	&0.182 	&0.529 &0.295	&1.509 &0.851 	 	&0.276 	&0.694 &0.423	&1.680 &0.846 	 	&0.276 	&0.691 &0.382	&1.789\\
&Deep-Net \cite{pan2016shallow}	                  &0.855 	 	&0.201 	&0.592 &0.331	&1.775 &0.884 	 	&0.300 	&0.736 &0.451	&2.066 &0.861 	 	&0.282 	&0.719 &0.414	&1.903\\
&DVA \cite{wang2017deep}                          &0.860  	&0.262  	 	&0.595  	&0.358  	&2.013 &0.886  	&0.372  	 	&0.727  	&0.482  	&2.459 &0.872  	&0.339  	 	&0.725  	&0.439  	&2.311 \\
\hline
\hline
Training&Ours                                   &0.885   &0.311       &0.553    &0.415   &2.259 &0.905   &0.471       &0.757    &0.577   &2.517 &0.894   &0.403       &0.742    &0.517   &2.559\\
setting  (i)&\textit{Attention module}       &0.854   &0.251       &0.545    &0.332   &1.755 &0.880   &0.415       &0.748    &0.529   &2.283 &0.853   &0.333       &0.719    &0.435   &1.946\\
\hline
Training&Ours                             &0.878   &0.297       &0.543    &0.388   &2.125 &0.912 &0.519 &0.754 &0.609 &3.049 &0.874   &0.364       &0.727    &0.452   &2.186\\
setting (ii)&\textit{Attention module}           &0.855   &0.250       &0.541    &0.318   &1.703 &0.885   &0.416       &0.690    &0.490   &2.113 &0.860   &0.322       &0.656    &0.367   &1.667\\
\hline
Training&Ours                          &0.866   &0.277       &0.596    &0.362   &1.951 &0.884   &0.449       &0.749    &0.534   &2.647 &\textbf{0.905}    &\textbf{0.496}    &\textbf{0.767}    &\textbf{0.603}    &\textbf{3.200}\\
setting (iii)&\textit{Attention module }            &0.852   &0.260       &0.582    &0.350   &1.945 &0.898 &0.429 &0.763 &0.543 &2.409 &0.884   &0.354       &0.743    &0.500   &2.339\\
\hline
Training &Ours                             &\textbf{0.890}   &\textbf{0.315}       &\textbf{0.601}    &\textbf{0.434}   &\textbf{2.354} &\textbf{0.913}   &\textbf{0.542}       &\textbf{0.757}    &\textbf{0.623}   &\textbf{3.086} &0.897   &0.406       &0.744    &0.510   &2.567\\
setting (iv) &\textit{Attention module}   &0.870   &0.273       &0.577    &0.380   &2.077  &0.878   &0.479       &0.686    &0.478   &2.060 &0.877   &0.379       &0.685    &0.411  &1.899\\
\hline
\end{tabular}
}
\begin{tablenotes}
\footnotesize
\item[]$^*$ Non-deep learning model.
\end{tablenotes}
\end{threeparttable}
\vspace*{0pt}
\caption{\textbf{Quantitative results on \ourdataset, Hollywood2, and UCF sports} datasets. The best scores are marked in \textbf{bold}. Training settings (\S\ref{sec:experimentsetup}) for video saliency datasets: (i) \ourdataset, (ii) Hollywood-2, (iii) UCF sports, and (iv) \ourdataset+Hollywood-2+UCF sports. }
\label{table4}
\vspace{-10pt}
\end{table*}

\noindent\textbf{Training/testing protocols.} We use the static stimuli ($10, 000$ images) from the training set of SALICON \cite{jiang2015salicon} dataset for training our attention module. For dynamic stimuli, we consider 4 settings: using the training set(s) from \textbf{(i)} \ourdataset, \textbf{(ii)} Hollywood-2, \textbf{(iii)} UCF sports, and \textbf{(iv)} \ourdataset+Hollywood-2+UCF sports. For \ourdataset, we use the original training/validation/testing splitting ($600/100/300$). For Hollywood-2, following \cite{marszalek2009actions}, $823$ videos for training and $884$ videos for testing. For UCF sports, the training and testing sets include $103$ and $47$ videos, respectively, as suggested by \cite{rodriguez2008action}. We randomly sample 10\% videos from the training sets of Hollywood-2, and UCF sports as their validation sets. We evaluate our model on the testing sets of \ourdataset, Hollywood-2, and UCF sports dataset, in total $1, 231$ video sequences with more than $400, 000$ frames.

\noindent\textbf{Implementation details.} Our model is implemented in Python on Keras, 
and trained with the Adam optimizer \cite{kingma2014adam}. During the training phase, the learning rate was set to $0.0001$ and was decreased by a factor of 10 every 2 epochs. The network was trained for 10 epochs. We perform early-stopping on the validation set.

\noindent\textbf{Competitors.} 
We compare our model with nine dynamic saliency models: PQFT \cite{guo2010novel}, Seo \textit{et al}. \cite{seo2009static}, Rudoy \textit{et al}. \cite{rudoy2013learning}, Hou \textit{et al}. \cite{hou2009dynamic},  Fang \textit{et al}. \cite{fang2014video},  OBDL \cite{hossein2015many}, AWS-D \cite{leboran2017dynamic}, OM-CNN \cite{jiang2017predicting}, and Two-stream \cite{bak2016two}\footnote{We re-implemented \cite{bak2016two} since
the official codes cannot run correctly.}. For the sake of complementary, we further compare with six state-of-the-art static attention models:  ITTI \cite{itti1998model}, GBVS \cite{harel2007graph}, SALICON \cite{huang2015salicon}, DVA \cite{wang2017deep}, Shallow-Net \cite{pan2016shallow}, and Deep-Net \cite{pan2016shallow}. OM-CNN, Two-stream, SALICON, DVA, Shallow-Net, and Deep-Net are deep learning models, and others are classical saliency models.  Those models are selected due to: 1) their representability of the diversity of the state-of-the-art; or 2) publicly available implementations. 

\noindent\textbf{Baselines.} We further derive $8$ baselines. For each training setting, we derive two baselines: \textit{Our} and \textit{Attention module}, refer to our final dynamic saliency prediction and the intermediate output of our attention module, respectively.


\noindent\textbf{Evaluation metrics.} Here, we employ five classic metrics, namely Normalized Scanpath Saliency (NSS), Similarity Metric (SIM), Linear Correlation Coefficient (CC), AUC-Judd (AUC-J), and shuffled AUC (s-AUC). Please refer to~\cite{borji2013state,wang2017deep} for detailed descriptions of these metrics.

\noindent\textbf{Computation load.} The whole model is trained in an end-to-end manner. The entire training procedure takes
about $30$ hours with a single NVIDIA TITAN X GPU and a 4.0GHz Intel processor (in training setting (iv)). Since our model does not need any pre- or post-processing, it takes only about $0.08$s to process an frame image of size $224\times224$.

\subsection{Performance comparison}
\label{sec:performance}
\vspace{-5pt}
\noindent\textbf{Performance on \ourdataset.}  Table \ref{table4} reports the comparative results with the
aforementioned saliency models, on the testing set ($300$ video sequences) of \ourdataset~dataset. In can be observed that the proposed model consistently and significantly outperforms other competitors, across all the metrics. This can be contributed to our specially designed attention module, which makes our model explicitly learn static and dynamic saliency representations in CNN and LSTM separately. Our model even does not use any optical flow algorithm (different with Fang \textit{et al}. \cite{fang2014video}, Two-stream \cite{bak2016two}). This significantly improves the applicability of our model and demonstrates the effectiveness of our training protocol that leveraging both static and dynamic stimuli.
\begin{figure*}
  \centering
      \includegraphics[width=\linewidth]{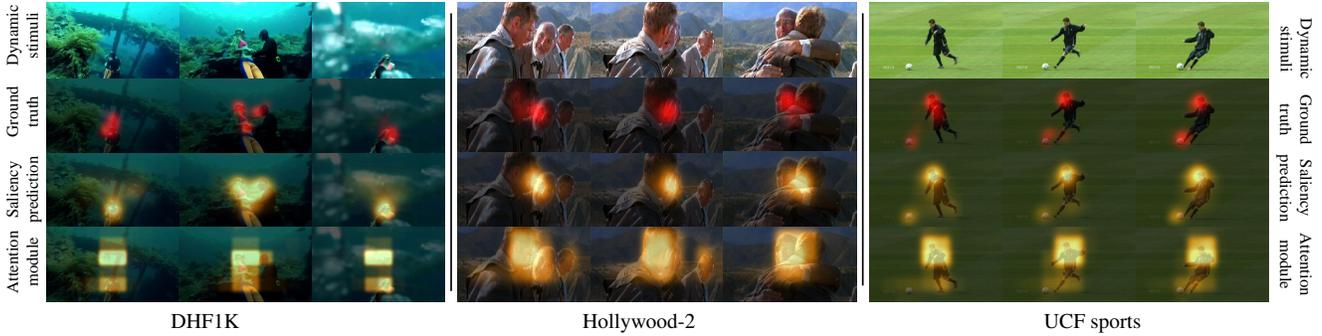}
           \\
     \mbox{}\hfill \footnotesize \ourdataset \hfill\mbox{}
     \mbox{}\hfill \footnotesize Hollywood-2 \hfill\mbox{}
     \mbox{}\hfill \footnotesize UCF sports \hfill\mbox{}
\caption{\textbf{Qualitative results} of our video saliency model on three datasets. Best viewed in color. }
\label{fig5}
\vspace{-10pt}
\end{figure*}

\noindent\textbf{Performance on Hollywood-2.} We further test our model on Hollywood-2 dataset, where the testing sets comprises $884$ video sequences. The results are summarized in Table \ref{table4}. Again, our model consistently significantly higher than
other methods across various metrics. Besides, when we go insight into the performance with training settings, the performance would increase with  increasing amount of training data. This suggests that the large-scale training data volume is important for the performance of neural network.

\noindent\textbf{Performance on UCF sports.} With the test set ($47$ video sequences) of UCF sports dataset, we again observe the proposed model provides consistently good results, compared to related state-of-the-art (see Table \ref{table4}). Interestingly, we find that, with small amount of training data (training setting (iii), $103$ video stimuli from UCF sports dataset), the proposed model achieves a very high performance, even better than the model (\textit{Our, training setting (iv)}) trained with large-scale data (1.5K video stimuli). This could be explained by lack of diversity in the video training data, as the videos in UCF sports dataset are highly related (with similar scenes and actors) and small scale. This is also consistent with our research for UCF sports which shows that $82.3$\% fixations are located on the human body area (see \S~\ref{sec:previousdatasets}).

\subsection{Analysis}
\vspace{-5pt}
\label{sec:analysis}
Based on our extensive experiments, we provide more detailed analyses, which would give deeper insights of previous studies and suggest some hints for future research.

\noindent\textbf{Dynamic saliency models: deep  \textit{vs} non-deep learning.} In dynamic scenes, previous deep learning based dynamic saliency models (\textit{i.e.}, OM-CNN, Two-stream) show significant improvements over classic dynamic models (\eg, PQFT, Seo \etal \textit{et al}., Rudoy \textit{et al}., Hou \textit{et al}., Fang \textit{et al}.). This demonstrates the strong learning ability of neural network and the promise of developing neural network in this challenging area.

\noindent\textbf{Non-deep learning models: static \textit{vs} dynamic.} An interesting finding is classic dynamic methods (\textit{i.e.}, PQFT, Seo \textit{et al}., Rudoy \textit{et al}., Hou \textit{et al}., Fang \textit{et al}.) did not perform better than their static counterparts: ITTI, GBVS. This is probably due to two reasons. First, the perceptual cues and underlying mechanisms of visual attention allocation during dynamic viewing are more complex and still not clear. Second, previous studies are more focused on computational models of static saliency, while less efforts were paid for modeling dynamic saliency.

\noindent\textbf{Deep learning models: static \textit{vs} dynamic.} Compared with state-of-the-art deep learning based static models (\textit{i.e.}, DVA, Deep-Net), previous deep learning based dynamic models (\textit{i.e.}, OM-CNN, Two-stream) only obtain slightly better performance (or only competitive).  Although strong motion information (\textit{i.e.}, optical flow, motion network) have been encoded into OM-CNN and Two-stream, their performance are still limited. We attribute this into the inherent difficulties of video saliency prediction and previous models' neglect of utilizing existing rich static saliency data.

\subsection{Ablation study}
\vspace{-5pt}
\label{sec:ablation}
In this section, we offer a more detailed exploration of our proposed approach in several aspects with \ourdataset~dataset. We verify the effectiveness of the proposed  mechanism, and examine the influence of different training protocols. The results are summarized in Table \ref{table7}.
\begin{table}[t]
\centering
\begin{threeparttable}
\resizebox{0.49\textwidth}{!}{
\setlength\tabcolsep{2pt}
\renewcommand\arraystretch{1.05}
\begin{tabular}{l|l|c|c|c|c|c}  
\hline\thickhline
Aspects &Variants &AUC-J$\uparrow$ &~SIM$\uparrow$~  &s-AUC$\uparrow$ &~~CC$\uparrow$~ &~NSS$\uparrow$~\\
\hline
\hline
Baseline&\tabincell{c}{\!\!\!\!\!\!training setting (iv) \\\footnotesize(\textit{1.5K videos+10K images})}                             &\textbf{0.890}   &\textbf{0.315}       &\textbf{0.601}    &\textbf{0.434}   &\textbf{2.354}\\
\hline
\hline
\multirow{3}*{Attention}&\tabincell{c}{w/o attention\\\footnotesize\!\!\!\!\!\!\!(\textit{1.5K videos})} &0.847 	&0.236 	 	&0.579 	&0.306 	&1.685\\
module&\tabincell{c}{w/o residual connection\\\footnotesize\!\!\!\!\!\!\!(\textit{1.5K videos+10K images})} &0.874 	&0.303 	 	&0.594 	&0.401 	&2.174\\
&\tabincell{c}{\!\!\!\!\!\!\!w/o downsampling\\\footnotesize(\textit{1.5K videos+10K images})} &0.870 	&0.298 	 	&0.583 	&0.389 	&2.085\\
\hline
\hline
\multirow{1}*{Training}&\tabincell{c}{reduced training samples\\\footnotesize\!\!\!\!\!\!\!\!\!\!\!\!\!\!(\textit{1.5K videos+5K images})}
&0.877  	&0.297  	 	&0.588  	&0.372  	&2.098\\
\hline
\hline
convLSTM&\tabincell{c}{\!\!\!\!\!\!\!\!\!\!\!\!\!w/o convLSTM \\\footnotesize(\textit{1.5K videos+10K images})}
&0.867 &0.269 &0.573 &0.382 &2.034\\

\hline
\end{tabular}
}
\end{threeparttable}
\vspace*{0pt}
\caption{\textbf{Ablation study on \ourdataset}. See \S\ref{sec:ablation} for details.}
\label{table7}
\vspace{-13pt}
\end{table}

\noindent\textbf{Effect of attention mechanism.} By disabling the attention module, and only training with video stimuli we observe a performance drop (\textit{e.g.}, AUC-J: $0.890$$\rightarrow$$0.847$), verifying the effectiveness of attention module and showing that the leverage of static stimuli indeed improves the predication accuracy in dynamic scenes. For exploring the effect of the residual connection in attention module (Eq. \ref{eq:3}), we train the model based on Eq. \ref{eq:2} (without residual connection). We observe a minor decrease; showing that employing residual connection could avoid distorting spatial features in frames.
In our attention module, we apply down-sampling for enlarging the receptive field. We also study the influence of such design. We find that the attention module with enlarged receptive field would gain better performance, since the model could make prediction in global view.

\noindent\textbf{Training.} We assess different training protocols. By reducing the amount of static training stimuli from 10K to 5K,
we observe a performance drop (\textit{e.g.}, AUC-J: $0.890$$\rightarrow$$0.877$). The baseline (\textit{w/o attention}) can also be viewed as the model without any static training stimuli, which gains worse performance  (\textit{e.g.}, AUC-J: $0.890$$\rightarrow$$0.847$). 


\noindent\textbf{Effect of convLSTM.} To study the influence of convLSTM, we re-train our model without convLSTM (using training setting (iv)) and get a baseline: \textit{w/o convLSTM}. We observe a drop of performance; showing that the dynamic information learnt in convLSTM could boost the performance.
\section{Discussion and Conclusion}
\vspace{-5pt}
\label{sec:conclusion}
In this work, we presented ``Dynamic Human Fixation (\ourdataset)'', a
large-scale carefully designed and systematically collected benchmark dataset to facilitate research in video saliency modeling. To the best of our knowledge, our work is the most comprehensive performance evaluation of video saliency models. \ourdataset~contains
1K videos, which capture representative instances, diverse contents and various motions, with human eye-tracking
annotations.

Further, we proposed a novel deep learning based video saliency model, which encodes a supervised attention mechanism to explicitly capture static saliency information and help LSTM better capture dynamic saliency representations over successive frames. We performed extensive experiments on \ourdataset, Hollywood-2, and UCF-sports datasets, and analyzed the performance of our model compared to previous attention models in dynamic scenes. Our experimental results demonstrate that our proposed model outperforms
other competitors and is quite efficient.



{\small
\bibliographystyle{ieee}
\bibliography{egbib}
}

\end{document}